\documentclass{article}

\usepackage[square,sort,comma,numbers]{natbib}
 \usepackage[final]{nips_2016}

\usepackage[utf8]{inputenc} 
\usepackage[T1]{fontenc}    
\usepackage{hyperref}       
\usepackage{url}            
\usepackage{booktabs}       
\usepackage{amsfonts}       
\usepackage{nicefrac}       
\usepackage{microtype}      
\usepackage{commath}
\usepackage{amsmath}
\usepackage{graphicx} 

\title{Ways of Conditioning Generative Adversarial Networks}

\author{
  Hanock Kwak and Byoung-Tak Zhang\\
  School of Computer Science and Engineering\\
  Seoul National University\\
  Seoul 151-744, Korea \\
  \texttt{\{hnkwak, btzhang\}@bi.snu.ac.kr} \\
}

\begin{document}

\maketitle

\begin{abstract}
  The GANs are generative models whose random samples realistically reflect natural images. It also can generate samples with specific attributes by concatenating a condition vector into the input, yet research on this field is not well studied. We propose novel methods of conditioning generative adversarial networks (GANs) that achieve state-of-the-art results on MNIST and CIFAR-10. We mainly introduce two models: an information retrieving model that extracts conditional information from the samples, and a spatial bilinear pooling model that forms bilinear features derived from the spatial cross product of an image and a condition vector. These methods significantly enhance log-likelihood of test data under the conditional distributions compared to the methods of concatenation.
  
\end{abstract}

\section{Introduction}

The goal of the generative model is to learn the underlying probabilistic distribution of unlabeled data by disentangling explanatory factors in the data\cite{bengio2009learning,theis2016a}. It is applicable to tasks such as classification, regression, visualization, and policy learning in reinforcement learning. The GAN\cite{goodfellow2014generative} is a prominent generative model that is found to be useful for realistic sample generation\cite{radford2015unsupervised}, semi-supervised learning\cite{salimans2016improved}, super resolution\cite{ledig2016photo}, text-to-image\cite{RAYLLS16}, and image inpainting\cite{pathakCVPR16context}. The potentiality of GAN was supported by the theory that if the model has enough capacity, the learned distribution can converge to the distribution over real data\cite{goodfellow2014generative}. The representative power of the GAN was highly enhanced with deep learning techniques\cite{radford2015unsupervised} and various methods\cite{zhao2016energy,salimans2016improved} was introduced to stabilize the learning process.

After the conditional GAN\cite{mirza2014conditional} was first introduced to generate samples of specific labels from a single generator, further research was not made thoroughly despite of its practical usage. For example, it was used to generate images from descriptive sentences\cite{RAYLLS16} or attribute vectors\cite{larsen2016autoencoding}. The conditioned GANs learn conditional probability distribution in which the condition can be any kind of auxiliary information describing the data. Mostly it was modeled by concatenating condition vectors into some layers of the generator and discriminator in GAN. Even though this forms a joint representation, it is hard to fully capture the complex associations between the two different modalities\cite{park2016multimodal}. We introduce a small variant of bilinear pooling that provides multiplicative interaction between all elements of two vectors. To provide computationally efficient and spatially sensible bilinear operation on an image (order 3 tensor) and a vector, we calculated cross product over last dimension of the image, resulting in a new image with increased channels.

If there is an oracle that can extract conditional information perfectly from any sample, we can train GAN to generate samples that the oracle can figure out given conditions from them. This is an information retrieval model where there is a pre-trained model that plays the role of an oracle. In addition to the objective of GAN, it maximizes lower bound of the mutual information between a given condition and the extracted condition. This idea has originated from infoGAN\cite{chen2016infogan} which is an information-theoretic extension to the GAN that is able to learn disentangled representations in a completely unsupervised manner.

\section{Related Works}

The conditional GAN\cite{mirza2014conditional} concatenates condition vector into the input of the generator and the discriminator. Variants of this method was successfully applied in \cite{RAYLLS16,larsen2016autoencoding,wang2016generative}. \cite{RAYLLS16} obtained visually-discriminative vector representation of text descriptions and then concatenated that vector into every layer of the discriminator and the noise vector of the generator. \cite{larsen2016autoencoding} used a similar method to generate face images from binary attribute vectors such as hair styles, face shapes, etc. In \cite{wang2016generative}, Structure-GAN generates surface normal maps and then they are concatenated into noise vector of Style-GAN to put styles in those maps. 

The spatial bilinear pooling was mainly inspired from studies on multimodal learning\cite{ngiam2011multimodal}. The key question of the multimodal learning is how can a model uncover the correlated interaction of two vectors from different domains. In order to achieve this, various methods (vector concatenation\cite{zhou2015simple}, element-wise operations\cite{yang2015stacked}, factorized restricted Boltzmann machine (RBM)\cite{ngiam2011multimodal}, bilinear pooling\cite{park2016multimodal,lin2015bilinear}, etc) were proposed for numerous challenging tasks. The RBM based models require expensive MCMC sampling which makes it difficult to scale them to large datasets. The bilinear pooling is more expressive then vector concatenation or element-wise operations, but they are inefficient due to squared complexity $O(n^2)$. To solve this problem, \cite{gao2016compact} addressed the space and time complexity of bilinear features using Tensor Sketch\cite{pham2013fast}.

The information retrieving model uses core algorithm of infoGAN\cite{chen2016infogan} that recover disentangled representations by maximizing the mutual information for inducing latent codes. In infoGAN, the input noise vector is decomposed into a source of incompressible noise and the latent code, and there is an auxiliary output in the discriminator to retrieve the latent codes. The infoGAN utilizes Variational Information Maximization\cite{agakov2004algorithm} to deal with intractability of the mutual information. Unlike infoGAN which randomly generates latent codes, we explicitly put condition information in the latent codes.

\section{Model}
\subsection{Generative Adversarial Networks}
The GAN\cite{goodfellow2014generative} has two networks: a generator $G$ that tries to generate real data given noise $z \sim p_{z}(z)$, and a discriminator $D \in [0,1]$ that classifies the real data $x \sim p_{data}(x)$ and the fake data $G(z)$. The $D(x)$ represents probability of $x$ being a real data. The objective of $G$ is to fit the true data distribution deceiving $D$ by playing the following minimax game:

\begin{equation}\label{eq1}
\begin{split}
\min_{\theta_{G}} \max_{\theta_{D}} \quad \mathbb{E}_{x \sim p_{data}(x)}[\log D(x)] + \mathbb{E}_{z \sim p_{z}(z)}[\log(1 - D(G(z)))],
\end{split}
\end{equation}

where $\theta_{G}$ and $\theta_{D}$ are parameters of $G$ and $D$, respectively. The parameters are updated by stochastic gradient descent (or ascent) algorithms.

\subsection{Conditioned GAN}
The objective of conditioned GANs is to fit conditional probability distribution $p(x|c)$ where $c$ is condition that describes $x$. The objective is to learn $p(x|c)$ correctly from a labeled dataset ${(x_1, c_1), (x_2, c_2), ..., (x_n, c_n)}$. The generator $G(z,c)$ of conditioned GAN has additional input $c$, and all generators used in our experiment take $[x,c]$ as an input where $[...]$ means vector concatenation. We can add $c$ into the input of  $D(x,c)$ as well, or put regularization terms to guide the generator. If we use $D(x,c)$, then the objective of the conditioned GAN is to optimize the equation \ref{eq1} for each $c$:

\begin{equation}\label{eq2}
\min_{\theta_{G}} \max_{\theta_{D}} \quad \mathbb{E}_{c \sim p_{data}(c)}[\mathbb{E}_{x \sim p_{data}(x|c)}[\log D(x,c)] + \mathbb{E}_{z \sim p_{z}(z)}[\log(1 - D(G(z,c), c))]],
\end{equation}

where $D(x,c)$ represents probability of $x$ being a real data from condition $c$. We can also train $G(x,c)$ to fit $p(x|c)$ by putting a regularization term $R(G)$ to the objective function:

\begin{equation}\label{eq3}
\min_{\theta_{G}} \max_{\theta_{D}} \quad \mathbb{E}_{x \sim p_{data}(x)}[\log D(x)] + \mathbb{E}_{z \sim p_{z}(z), c \sim p_{data}(c)}[\log(1 - D(G(z,c)))] + R(G).
\end{equation}

\subsection{Concatenation Methods}

We compare our proposed models with two commonly used conditioned GANs. The first one is a conditional GAN\cite{mirza2014conditional} (CGAN) that concatenates $c$ into $x$ of $D$, and the second one is a fully conditional GAN (FCGAN) that concatenates $c$ into every layer of $D$ including $x$. When the dimension of a layer in $D$ is $n\times n\times d$ where $d$ is the size of depth, we replicate the vector $c$ spatially to match the size $n\times n$ of the feature map and perform a depth concatenation. Both models, including ours, use the same structure for $G(z,c)$ where $c$ is only concatenated to $z$. 

\subsection{Spatial Bilinear Pooling}

We propose Spatial Bilinear Pooling (SBP) that provides multiplicative interaction between all elements of two vectors. When the dimension of an image is $n\times n\times d$, the SBP performs cross product for each pixel ($1\times 1\times d$) of the image with $c$ and then gathers the resulting vectors spatially to make a new image.

\subsection{Information Retrieving GAN}

The Information Retrieving GAN (IRGAN) has an approximator $Q(c|x)$ that measures $p(c|x)$ from $x$. The generator $G(z,c)$ of IRGAN maximizes $Q(c|G(z,c))$, fitting the conditional distribution $p(x|c)$. In our experiments, the approximators are pre-trained classifiers for MNIST and CIFAR-10. IRGAN optimizes the equation \ref{eq3} where $R(G)$ is defined by the lower bound\cite{chen2016infogan} of the mutual information $I(c;G(z,c))$:

\begin{equation}\label{eq4}
\begin{split}
R(G) = -\lambda \mathbb{E}_{c \sim p_{data}(c), x \sim G(z,c)} [\log Q(c|x)].
\end{split}
\end{equation}

In this paper we assume that the entropy $H(c)$ is constant and omit it from $R(G)$ for simplicity. The only difference from \cite{chen2016infogan} is that $c$ is sampled from the data distribution rather than a pre-defined distribution, since $c$ are explicitly given from the datasets.

\begin{figure*}[t]
	\vskip 0.0in
	\begin{center}
		\includegraphics[width=\textwidth]{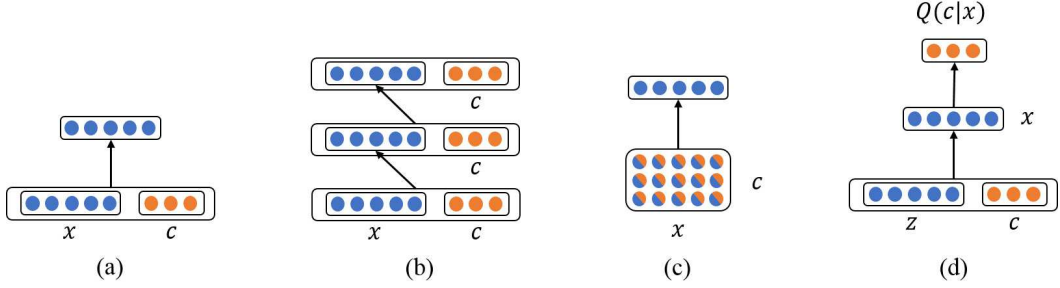}
		\caption{Illustration of the four conditioned GANs: (a) CGAN, (b) FCGAN, (c) SBP, and (d) IRGAN. For simplicity of visualization, only a single pixel of $x$ is visualized, virtually consisting of 5 channels. }
		\label{model_fig}
	\end{center}
	\vskip 0.0in
\end{figure*} 

\begin{table}[t]
  \caption{Parzen window-based log-likelihood estimates on MNIST.}
  \label{mnist_table}
  \def\arraystretch{1.2}
  \centering
  \begin{tabular}{ccccccccccc}
		\hline
		\multicolumn{1}{c|}{\textbf{}}   & \multicolumn{10}{c}{\textbf{Label of MNIST}}   \\ \hline
		\multicolumn{1}{c|}{\textbf{Model}}   & \textbf{0} & \textbf{1} & \textbf{2} & \textbf{3} & \textbf{4} & \textbf{5} & \textbf{6} & \textbf{7} & \textbf{8} & \textbf{9}   \\ \hline
		\multicolumn{1}{c|}{\textbf{CGAN}}  & 124.7  & 358.0  & 35.7  & 56.9  & 106.7  & -10.6  & 113.7  & 183.5  & 53.1  & 166.2    \\ 
		\multicolumn{1}{c|}{\textbf{FCGAN}} & 28.7  & 531.4  & -11.9  & 52.8  & 101.1   & 51.7  & 81.6  & 221.0  & 20.0  & 187.1   \\
		\multicolumn{1}{c|}{\textbf{SBP}}   & 153.0  & 643.7  & \textbf{81.6}  & 124.6  & 173.2   & 121.4  & 197.2  & 287.5  & 83.1  & 248.7   \\
		\multicolumn{1}{c|}{\textbf{IRGAN}} & \textbf{154.0}  & \textbf{749.7}  & 68.7  & \textbf{137.3}  & \textbf{208.8}   & \textbf{130.8}  & \textbf{214.8}  & \textbf{296.5}  & \textbf{97.7}  & \textbf{268.4}   \\ \hline
  \end{tabular}
  \vspace{0.5cm}
  \caption{Parzen window-based log-likelihood estimates on CIFAR-10.}
  \label{cifar10_table}
  \begin{tabular}{ccccccccccc}
		\hline
		\multicolumn{1}{c|}{\textbf{}}   & \multicolumn{10}{c}{\textbf{Label of CIFAR-10}}   \\ \hline
		\multicolumn{1}{c|}{\textbf{Model}}   & \textbf{airplane} & \textbf{car} & \textbf{bird} & \textbf{cat} & \textbf{deer} & \textbf{dog} & \textbf{frog} & \textbf{horse} & \textbf{ship} & \textbf{truck}   \\ \hline
		\multicolumn{1}{c|}{\textbf{CGAN}}  & 684.0  & 417.4  & 969.5  & 603.8  & 1064.5  & 547.8  & 929.9  & 554.9  & 771.1  & 373.5    \\ 
		\multicolumn{1}{c|}{\textbf{FCGAN}} & 760.2  & 384.7  & 1018.9  & 566.3  & 1099.5   & 538.5  & \textbf{985.9}  & 558.5  & 793.2  & 383.7   \\
		\multicolumn{1}{c|}{\textbf{SBP}}   & \textbf{847.7}  & \textbf{471.3}  & \textbf{1057.2}  & \textbf{625.9}  & \textbf{1180.9}   & 577.1  & 974.7  & \textbf{606.8}  & \textbf{872.9}  & \textbf{460.6}   \\
		\multicolumn{1}{c|}{\textbf{IRGAN}} & 721.8  & 391.1  & 1038.0  & 561.1  & 1027.9   & \textbf{586.4}  & 831.4  & 509.0  & 744.4  & 365.9   \\ \hline
  \end{tabular}
\end{table}

\section{Experiments}

We trained conditioned GANs on MNIST and CIFAR-10, and utilized the techniques proposed by DCGAN\cite{radford2015unsupervised}. All generators have identical structure, and the variations only exist in the discriminators and the auxiliary networks. We measured log-likelihood of the test set data for each $c$ by fitting a Gaussian Parzen window to the samples generated from $G(z,c)$\cite{goodfellow2014generative}. The best standard deviation $\sigma$ of the Gaussian was chosen by the validation set. 

The results are shown in Table \ref{mnist_table}, \ref{cifar10_table}. The IRGAN showed the best result on MNIST, but it didn't performed well on CIFAR-10 due to inaccurate classifier $Q(c|x)$ whose classification accuracy was 81\%. We found that SBP showed stable results on both datasets and most of the labels, surpassing CGAN and FCGAN.  

\section{Conclusions and Future Works}
 
In this paper we proposed effective models that can fit conditional distribution $p(x|c)$ more accurately. However, we need more experiments on other complex conditions such as multi-labels, text descriptions, and style embedding to verify our models. Since cross product of two long vectors is inefficient, compressing bilinear pooling is a promising alternative to the spatial bilinear pooling when conditions are high dimensional vectors. 

\bibliographystyle{unsrt}
\bibliography{nips_2016}

\end{document}